\documentclass[letterpaper, 10 pt, conference]{ieeeconf}
\IEEEoverridecommandlockouts                              

\usepackage{cite}
\usepackage{multirow}
\usepackage{algorithm}  	
\usepackage{algorithmic}
\usepackage{graphicx}
\usepackage{times}
\usepackage{amsmath}
\usepackage{amssymb}
\usepackage{subfigure}
\usepackage{graphics} 
\usepackage[utf8]{inputenc}
\usepackage{boldline}
\usepackage{multicol}
\usepackage[pdfstartview=FitH,bookmarksnumbered=true,colorlinks,bookmarksopen=true]{hyperref}

\title{\LARGE \bf
PointNetGPD: Detecting Grasp Configurations from Point Sets 
}

\author{Hongzhuo Liang$^{1\dagger}$, Xiaojian Ma$^{2\dagger}$, Shuang Li$^{1}$, Michael G\"{o}rner$^{1}$, Song Tang$^{1}$, \\Bin Fang$^{2}$, Fuchun Sun$^{2*}$, Jianwei Zhang$^{1}$
\thanks{$\dagger$These two authors contributed equally. This work was done when Hongzhuo Liang was visiting Tsinghua University.}
\thanks{$^{1}$TAMS (Technical Aspects of Multimodal Systems), Department of Informatics, Universit\"{a}t Hamburg}
\thanks{$^{2}$Tsinghua National Laboratory for Information Science and Technology (TNList), State Key Lab on Intelligent Technology and Systems, Department of Computer Science and Technology, Tsinghua University}
\thanks{*Corresponding author to provide e-mail: fcsun@tsinghua.edu.cn}%
}
\begin{document}

\maketitle
\thispagestyle{empty}
\pagestyle{empty}

\begin{abstract}
In this paper, we propose an end-to-end grasp evaluation model to address the challenging problem of localizing robot grasp configurations directly from the point cloud. Compared to recent grasp evaluation metrics that are based on handcrafted depth features and a convolutional neural network (CNN), our proposed PointNetGPD is lightweight and can directly process the 3D point cloud that locates within the gripper for grasp evaluation. Taking the raw point cloud as input, our proposed grasp evaluation network can capture the complex geometric structure of the contact area between the gripper and the object even if the point cloud is very sparse. To further improve our proposed model, we generate a larger-scale grasp dataset with 350k real point cloud and grasps with the YCB object set for training. The performance of the proposed model is quantitatively measured both in simulation and on robotic hardware. Experiments on object grasping and clutter removal show that our proposed model generalizes well to novel objects and outperforms state-of-the-art methods. Code and video are available at \href{https://lianghongzhuo.github.io/PointNetGPD}{https://lianghongzhuo.github.io/PointNetGPD}.
\end{abstract}

\section{INTRODUCTION}
Planning a grasp under uncertainty is a difficult task in robotics. For a robot that operates in the real world, uncertainty may come from varied aspects. In this paper, we mainly concentrate on the uncertainty brought by the imprecision and deficiency in sensing. This kind of uncertainty is usually associated with the sensor we use for robotic perception~\cite{varley2015generating}. To address this problem, a grasping model that can work with raw sensor input is needed. Some recent advances suggest to use deep neural networks that have been trained on large-scale grasp dataset labeling by humans~\cite{dex-net1,dex-net2} or grasping outcomes done by robotic hardware~\cite{guo2016object, guo2017robotic} to plan grasps directly with sensor input like images~\cite{lenz2015deep} or point cloud~\cite{gpg}. Such research work yields promising results across a wide variety of objects, sensors, and robots, and their models generalize well to novel objects that are not present in the training set. However, most of the current methods still rely on 2D (image) or 2.5D (depth map) input; some grasping models even require complex hand-crafted features~\cite{gpd} before they can process the data, while very few of them will take the 3D geometry information into consideration~\cite{yan2017learning}. Intuitively, whether a grasp is successful or not is always related to how the robot (gripper) interacts with the object surface in 3D space; thus the lack of geometry analysis could entail side effects to grasp planning, especially when accurate and complete sensing is not available.

\begin{figure}[t!]
    \centering
    \includegraphics[width=0.45\textwidth]{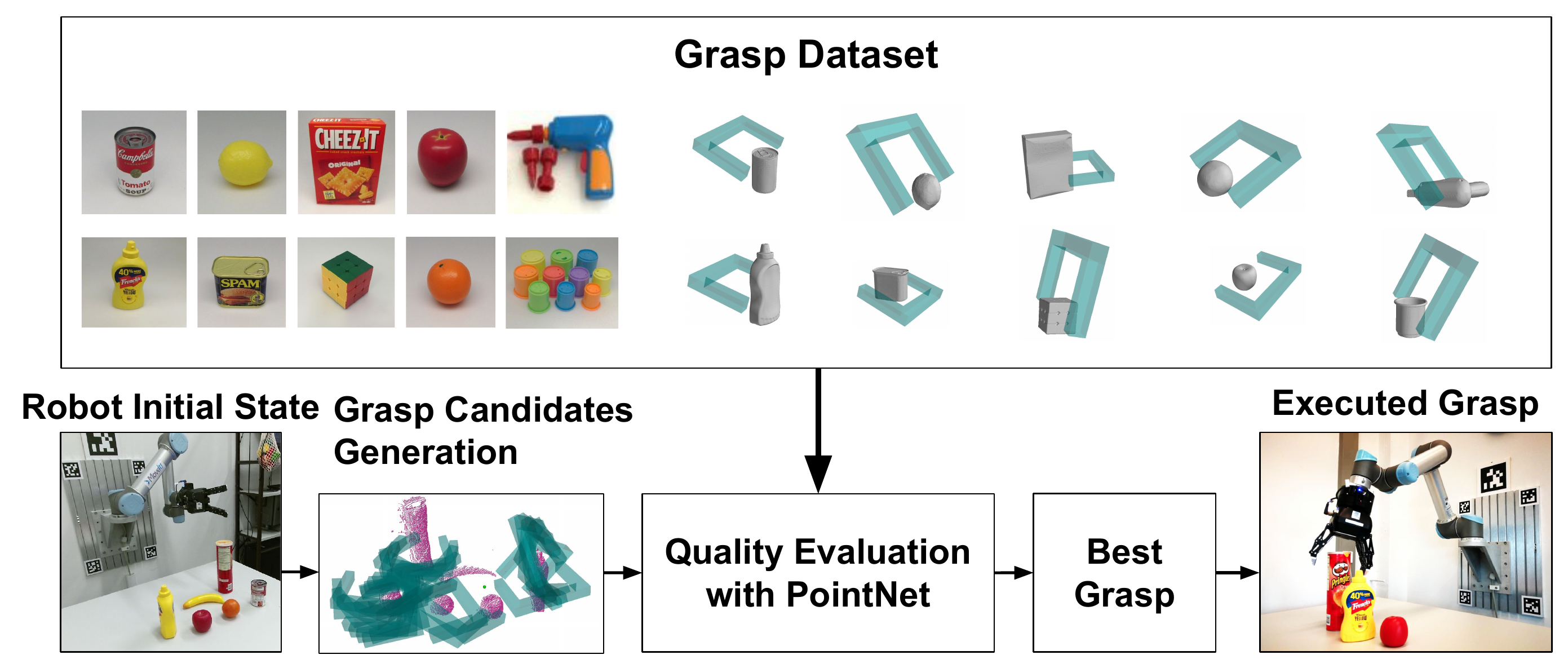}
    \caption{An illustration of our proposed PointNetGPD for detecting reliable grasp configuration from point sets. Taking raw sensor input from a common RGB-D camera, we first convert the depth map into a point cloud, then several grasp candidates will be sampled with essential geometry information as heuristic or constraints. For each candidate, the point cloud within the gripper will be cropped and transformed into local coordinates and finally fed into our grasp quality evaluation network. The grasp with the highest score will be executed. Our model is trained with a large-scale grasp dataset based on the YCB~\cite{YCB} object set.}
    \label{fig:pipeline}
\end{figure}

To tackle these unsolved issues, inspired by the recent work of PointNet~\cite{pointnet} that directly operates on point sets for 3D object classification and segmentation, in this work, we propose a point cloud based grasp detection method for detecting reliable grasp configurations from the point cloud. As illustrated in Figure~\ref{fig:pipeline}, PointNetGPD provides an effective pipeline to generate and evaluate grasp configurations. Compared with previous grasp detection methods that depend on multi-view CNN~\cite{gpd} or 3D-CNN~\cite{varley2017shape}, our approach does not require point cloud projection on multiple 2D images or rasterization into dense 3D volumes. As a result, it could mostly sustain the geometric information of the original point cloud and infer grasp quality more efficiently.

Recent success in deep neural network based grasp detection methods~\cite{dex-net2,lenz2015deep} emphasizes the importance of training on large-scale datasets. To further improve the performance of the proposed grasp detection method, we built a grasp dataset with a 350k real point cloud captured by depth cameras, parallel-jaw grasps and analytic grasp metrics over a subset of the YCB~\cite{YCB} object set. Different from other grasp datasets like Dex-Net~\cite{dex-net2}, we provide fine-grained scores for each grasp instead of binary labels. Specifically, given a 6D grasp pose and CAD model of an object, we perform force-closure~\cite{antipodal} and a friction-less grasp wrench space (GWS)~\cite{gws} analysis on the grasp respectively to obtain such scores. Quantitative scores make the more flexible label assignment possible during training, which could also improve the performance of our grasp quality evaluation network.

To summarize, our key contributions are twofold:
\begin{itemize}
    \item We propose to evaluate the grasp quality by performing geometry analysis directly from a 3D point cloud based on the network architecture of PointNet~\cite{pointnet}. Compared with other CNN-based methods~\cite{dex-net2,gpd,yan2017learning}, our method can exploit the 3D geometry information in the depth image better without any hand-crafted features and sustain a relatively small amount of parameters for learning and inference efficiency. Also, we found our proposed method still works well even when the point cloud is very sparse, which implies its potential for planning grasps under imprecise and deficient sensing.  
    \item We built a large-scale grasp dataset that contains 350k real point cloud and parallel-jaw grasps. Meticulous grasp quality scores that combine the force-closure and GWS analysis are provided. Our experiments demonstrate that our grasping model can obtain significant performance promotion from these meticulous scores and labels.    
\end{itemize}

\section{RELATED WORK}
\noindent
\textbf{Grasp Configuration Detection.}
Given an object (or a clutter) and essential environmental constraints, the goal of grasp configuration detection is to find a gripper configuration that maximizes the grasp quality metrics. Existing methods on this problem usually fall into one of two categories: model-based or model-free. Model-based approaches~\cite{zeng2016multi,dex-net1} typically rely on a pre-built grasp database of common 3D object models labeled with sets of feasible grasps and quality metrics provided by assisted tools like GraspIt!~\cite{graspit}. During execution, they need to associate the sensor input with an object entry in the database for grasp planning. Such matching is mainly based on visual and geometrical similarity~\cite{bohg2014data,brook2011collaborative,hinterstoisser2011multimodal}. However, due to imprecise sensing and due to the limited size of the database, model-based methods could arguably have poor generalized performances on novel objects and objects that are presented in dense clutter. 
In contrast, model-free methods are usually composed of two separate parts: grasp candidate generation and grasp quality metrics. In the first part, the geometry information captured by sensors will be leveraged as a heuristic or constraint~\cite{gpg} to build an adaptive grasp configuration sampler over the given object. These grasp candidates will then be evaluated by the quality metrics. In some modern model-free methods, large grasp datasets are also needed for training better quality metrics based on deep neural networks~\cite{dex-net2,guo2016object}.     

\noindent
\textbf{Grasp Quality Metrics.}
To evaluate the quality of a grasp, many analytic approaches physically analyze the geometry of the gripper configuration and the object to evaluate the quality of a grasp. Force-closure~\cite{antipodal,force-closure} and GWS analysis are two mainstream grasp quality metrics. Force-closure methods take the friction between the gripper and object into consideration, while GWS could work on friction-less cases. However, these analytic methods can provide reliable grasp quality measurements only when the precise object model is available; thus they cannot handle raw sensor input like the point cloud.

\noindent
\textbf{3D Computer Vision in Robotic Grasping.}
For robotic grasping, one of the challenges is posed by the uncertainty of perception. Since the robotic gripper needs to interact with the object in 3D space, precise and finer 3D visual analysis will be critical for a successful grasp. Motivated by the success of deep neural networks in various 3D computer vision tasks~\cite{mv3d,dss,voxelnet,sonet}, several trials on combining 3D computer vision techniques and grasp planning have been carried out~\cite{gpd,dex-net2,guo2016object,varley2017shape,yan2017learning}. 

In \cite{zeng2016multi}, researchers introduced a pipeline for grasping objects from dense clutter. They utilize multi-view depth input to eliminate deficiencies in depth sensing, but their proposed method strongly depends on accurate CAD models of the grasped objects, which will be impossible for generalizing to novel objects. 

The authors of \cite{varley2017shape} proposed to conduct convolution on a voxelized 3D grid (3D-CNN) from a point cloud to obtain the geometry representation of grasping objects. This representation will then be fed into a grasp generation model. Inferring a grasp with 3D-CNN could improve the analysis of grasp geometry. However, one of the main drawbacks of this method is that the runtime and memory complexity grows cubically with the resolution of the input 3D voxels~\cite{octnet}. As a result, the input will have to be limited to a pretty low resolution. Furthermore, the sparsity of the point cloud may even distract the neural network from learning meaningful features of grasp geometry since most of the voxels will not be occupied by any points. 

GPD~\cite{gpd} is the work closest to ours. The authors designed several projection features on normalized point cloud to construct a CNN-based grasp quality evaluation model and reach state-of-the-art performance in grasping objects from dense clutter. However, due to the network architecture and hand-crafted depth features, in our experiments, we found that GPD suffers from severe overfitting and performance reduction when the input point cloud is overall sparse (results and analysis can be found in Section~\ref{sec:sim_exp}). On the other hand, in most real-world grasping situations, it could be hard to obtain a relatively comprehensive point cloud especially when the clutter is highly occluded.

In conclusion, the approaches mentioned above mostly use 2D or 2.5D input, which has been found to be insufficient for geometry analysis. By introducing PointNet~\cite{pointnet} for 3D representation learning and meticulous grasp quality labels for supervision, our proposed method can outperform these results regarding both grasping performance and efficiency.

\section{Problem Formulation}
\subsection{Definitions}
Given a specified object $o$, things that are related to grasping will be the coefficient of friction between the object and gripper $\gamma \in \mathbb{R}$, the object's geometry and mass properties $M_o$,  and 6 DOF pose $W_o \in \mathbb{R}^6$. Let $\mathbf{s}_o = (W_o, M_o, \gamma)$ represent the state of the object. We denote a grasp configuration in 3D space as $\mathbf{g} = (\mathbf{p}, \mathbf{r}) \in \mathbb{R}^6$, where $\mathbf{p}  =(x,y,z) \in \mathbb{R}^3$ and $\mathbf{r} = (r_x, r_y, r_z) \in \mathbb{R}^3$ specify the position and orientation of the gripper respectively.  We only consider parallel-jaw grippers in this paper. Also, we assume a camera to capture the depth map, and the converted point cloud that contains N points is denoted as $\mathbf{P} \in \mathbb{R}^{3 \times N}$. For simplicity, all spatial quantities are in camera coordinates. To evaluate the quality of a grasp, we denote a quality metric as $Q(\mathbf{s}, \mathbf{g}) \rightarrow \mathbb{R}$. Notice that $Q$ works with an accurate object state instead of a point cloud, and our grasp quality is a continuous quantity instead of a binary label. 
\subsection{Objective}\label{sec:problem}
Given a gripper configuration $\mathbf{g}$ and sensor observation $\mathbf{P}$, our goal is to learn a quality metric $Q_\theta(\mathbf{P}, \mathbf{g}) \in \{c_0, c_1, \cdots\}$ to 
predict grasp quality from a point cloud. $\theta$ defines the parameters of our proposed grasp quality evaluation network described in Section~\ref{sec:pointgpd}. $c_0, c_1, \cdots$ are labels 
that represent the quality of a grasp $\mathbf{g}$, and can be assigned to any ground truth quality metrics $Q(\mathbf{s}, \mathbf{g})$. 

\section{End-to-End Learning Of Grasp Quality Metrics}
There are two main challenges to solving the problem in Section~\ref{sec:problem}.
First, learning such a grasp quality metric may require a massive number of samples over a wide range of objects to achieve good performance and generalization. Second, the input point cloud $\mathbf{P}$ could be imprecise and deficient, which leads to additional difficulties in geometry analysis. Consequently, we propose to evaluate the grasp quality by direct point cloud analysis with PointNet~\cite{pointnet}, and train our grasp quality metric on a generated large-scale dataset of 350k real point cloud and grasps over objects from the YCB~\cite{YCB} object sets to obtain robust grasp classification results.

\subsection{A Grasp Dataset with Meticulous Scores}\label{sec:dataset}
The generation of our grasp dataset involves two steps: sampling and scoring. Grasp candidates are firstly sampled over provided object meshes; then these candidates will be labeled by robust grasp quality metrics including force-closure and GWS, details are listed as follows:

\noindent
\textbf{Sampling.}
Although the YCB~\cite{YCB} provides registered point cloud for most of the objects, we still sample over the precise meshes instead to prevent the sampler from generating unfeasible grasps (such as grasps that collide with the object). For each grasp, we randomly sample two surface points $\mathbf{p}_1$, $\mathbf{p}_2$ as contact points and an approach angle between $[0, 0.5\pi)$ 
then a grasp $\mathbf{g}((\mathbf{p}_1 + \mathbf{p}_2)/2, \mathbf{r})$ will be constructed. To further eliminate unfeasible grasps, we conduct a sanity check by simulating the approach and close-finger action with a gripper model to see whether it will collide with the object. Finally, all the remaining grasps would then be transformed from mesh into point cloud coordinates. The transform matrices are obtained by doing ICP between the mesh and corresponding registered point cloud.

\begin{figure}[H]
    \centering
    \subfigure[]{\includegraphics[height=0.15\textwidth]{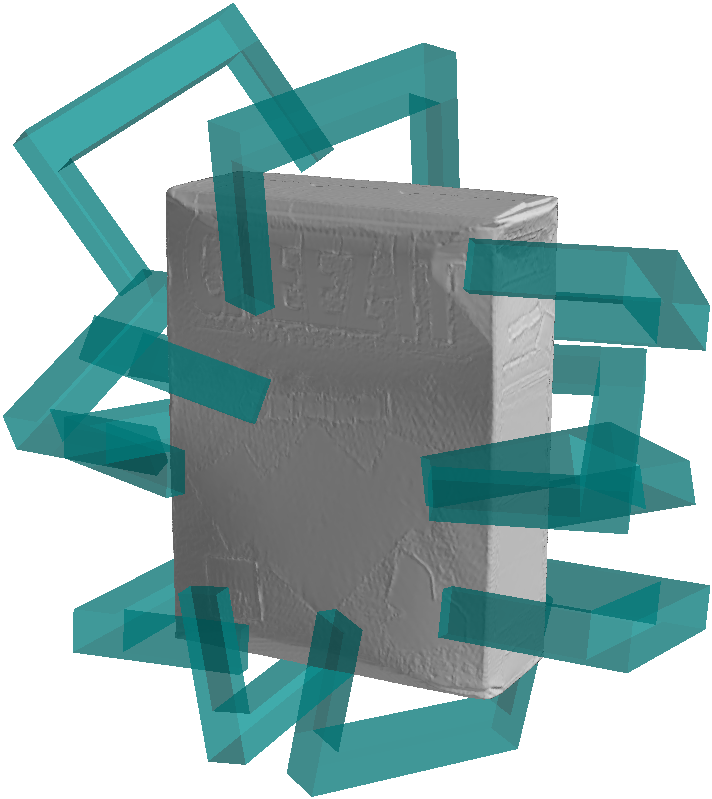}
        \label{fig:0.4}}
    \hspace{0.4in}
    \subfigure[]{\includegraphics[height=0.15\textwidth]{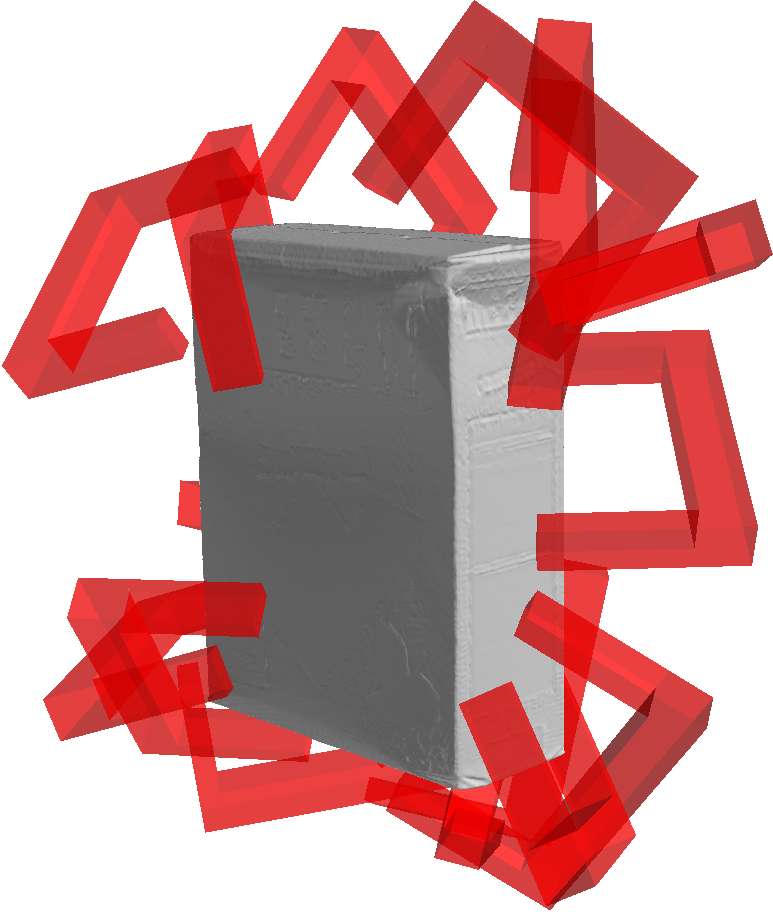}
        \label{fig:2.0}}
    \caption{Example grasps in our dataset label with $Q_{fc}$. \subref{fig:0.4} green grasps are labeled with $\gamma = 0.4$. \subref{fig:2.0} red grasps are labeled with $\gamma = 2.0$. We found that on this relatively simple box-like object, there is a significant difference in robustness between the green and red grasps.}
    \label{fig:dataset_example}
\end{figure}

\noindent
\textbf{Scoring.}
Given a sampled grasp $\mathbf{g}$ and object state $\mathbf{s}$, we adopt two different robust grasp quality metrics to label the grasp. 
One of them is a force-closure metric $Q_{fc}$; it requires the coefficient of friction $\gamma$ and only provides a binary outcome that indicates whether the grasp is force-closure or not. Here we modify it to enable quantitative scoring: Starting from $0.4$, we gradually increase $\gamma$ until the grasp is antipodal, then the value $1/\gamma$ will be recorded as a score for the current grasp. Such modification is intuitive since an antipodal grasp that requires lower friction could be arguably better. As is shown in Figure~\ref{fig:dataset_example}, the grasp with lower $\gamma$ could be more robust and feasible. We also observe that such difference will be more notable when the object has a more complex physical shape.  
The other grasp metric $Q_{gws}$ is based on Grasp Wrench Space (GWS) analysis~\cite{gws}. Compared to $Q_{fc}$, GWS analysis proposes to use the radius of GWS as a quantitative score of grasp quality. GWS itself can either be a $\mathbb{R}^3$ or $\mathbb{R}^6$ space. In practice, here we only apply a simplified $Q_{gws}$ with $\mathbb{R}^3$ friction-less grasp wrench space. 

We adopt a weighted sum to combine these two kinds of metrics, and produce a final quality score:
\begin{align}\label{label_score}
    Q(\mathbf{s}, \mathbf{g}) = \alpha Q_{fc}(\mathbf{s}, \mathbf{g}) + \beta Q_{gws}(\mathbf{s}, \mathbf{g})\text{.}
\end{align}
We observe that $Q_{gws}$ could be much larger than $Q_{fc}$ for most of grasps and objects, thus we choose $(\alpha, \beta) = (1.0, 0.01)$ in our experiments.

\begin{figure*}
    \centering
    \includegraphics[width=1.0\textwidth]{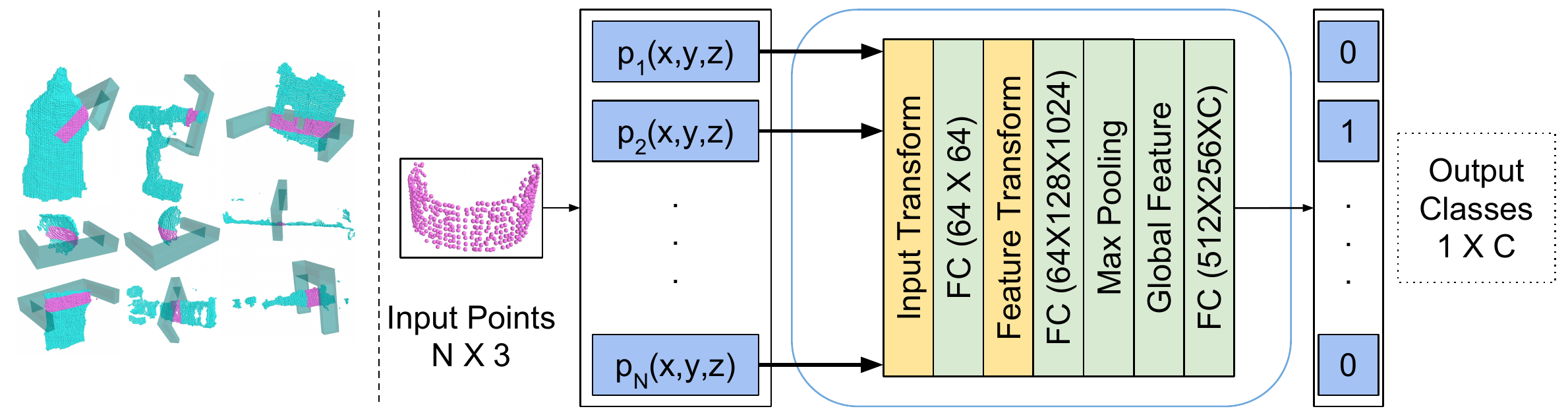}
    \caption{Architecture of our grasp quality evaluation network based on PointNet~\cite{pointnet}. Given a grasp and point cloud, the grasp is represented by the points within the closing area of the gripper. As is shown in Figure~\ref{fig:local_coord}, all the points will be transformed into local gripper coordinates before fed into the network. After several spatial transformations and feature extractions, the final global feature will be applied to classify the quality level of the input grasp.}
    \label{fig:pointnet}
\end{figure*}

\subsection{Learning a Grasp Quality Metric from Point Cloud}\label{sec:pointgpd}
\begin{figure}[H]
    \centering
    \subfigure[]{\includegraphics[height=0.15\textwidth]{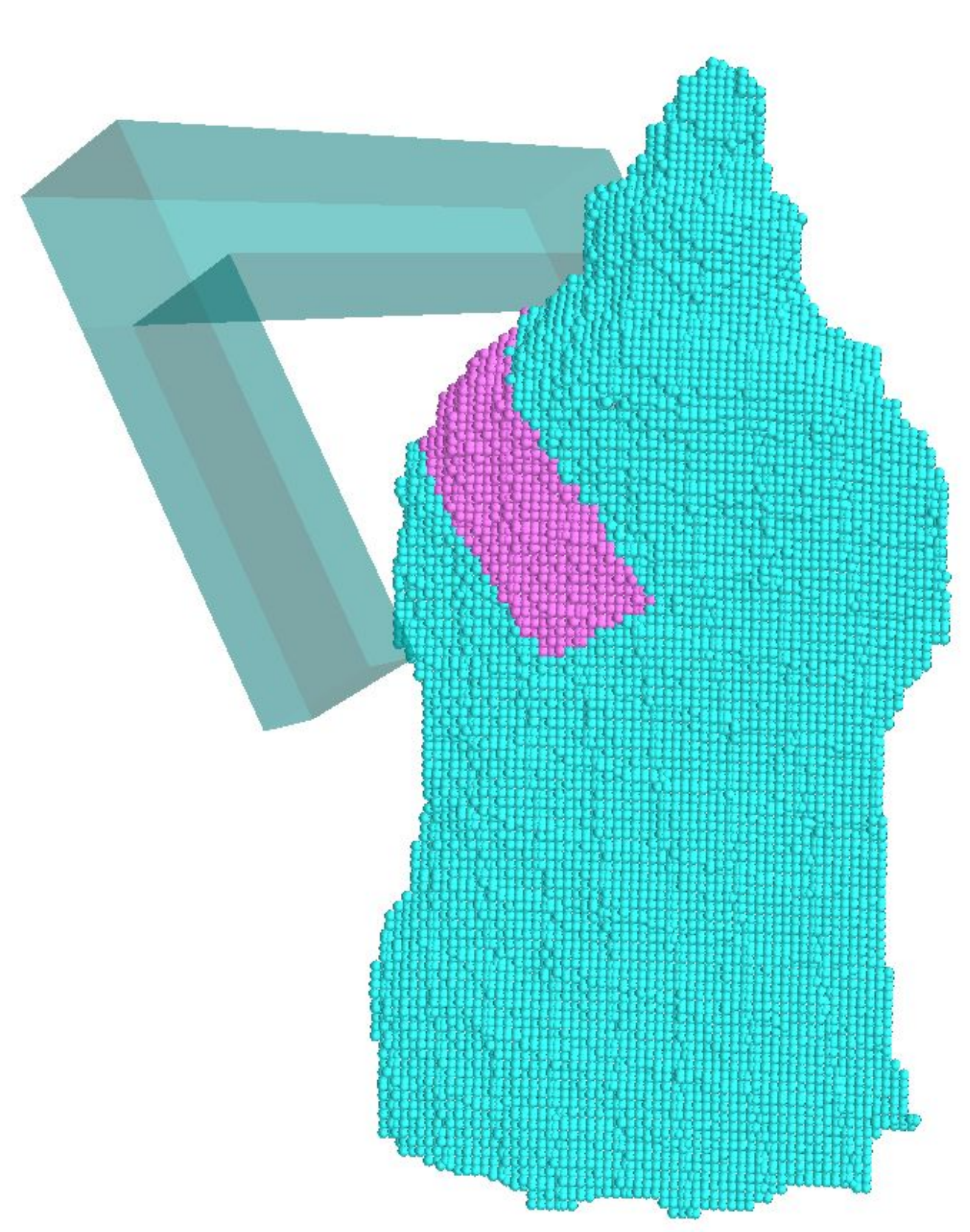}
        \label{fig:grasp_example}}
    \hspace{0.1in}
    \subfigure[]{\includegraphics[height=0.15\textwidth]{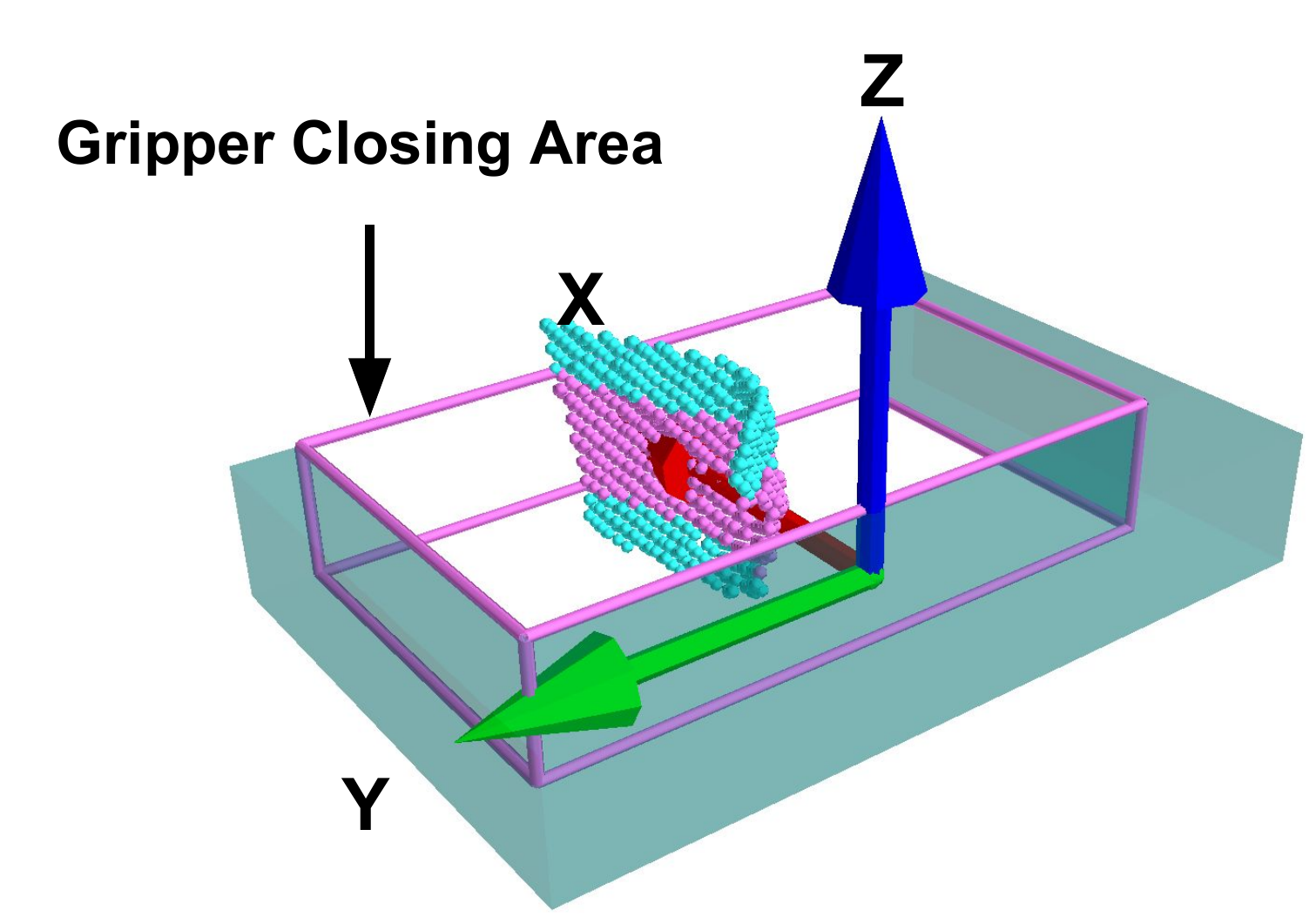}
        \label{fig:grasp_representation}}
    \caption{Grasp representation in local gripper coordinates. A grasp is represented by the point cloud within the gripper closing area. \subref{fig:grasp_example} a typical grasp configuration. \subref{fig:grasp_representation} axes of local coordinates and the transformed point cloud within the gripper closing area (magenta) that serves as the grasp representation.}
    \label{fig:local_coord}
\end{figure}

\noindent
\textbf{Network Architecture and Grasp Representation.} 
The architecture of our grasp quality evaluation network is illustrated in Figure~\ref{fig:pointnet}. Our PointNet~\cite{pointnet} like network will take as input the grasp represented by the point cloud within the closing area of the gripper. For learning and inference efficiency, we do not take the whole point cloud as input like~\cite{dex-net2,lenz2015deep}. The point cloud will firstly be transformed into the unified local gripper coordinate introduced in Figure~\ref{fig:local_coord}, this is mainly to eliminate the ambiguity caused by the different experiment (especially camera) settings. Specifically, we treat the approaching, parallel and orthogonal directions of the gripper as the XYZ axes respectively, while the origin will be located at the bottom center of the gripper. Then these $N\time3$ points will be passed through the network to estimate the level of quality. Compared to other CNN-based grasp quality evaluation networks, our model is lightweight and only has approximately 1.6 million parameters.

\noindent
\textbf{Training Dataset.}
We use the dataset generated in section~\ref{sec:dataset} to train our grasp quality evaluation model. Since there are quantitative quality~\eqref{label_score} values instead of binary labels in our dataset, it will be flexible to assign classifying labels and even enable the multi-class grasp quality classification. The threshold for each label will be discussed in Section~\ref{sec:sim_exp}. There are 350k point cloud and grasps over 47 YCB objects. To keep a balance of grasps with different qualities, we sample an equal number of grasps with $Q_{fc}$ value from $\{1/0.4, 1/0.45, 1/0.5, 1/0.8, 1/1.2, 1/1.6, 1/2.0\}$. For the point cloud, as suggested in~\cite{gpd}, we use real point cloud provided by YCB instead of simulated point cloud obtained with a CAD model for a better generalization to real world grasping tasks.

\noindent
\textbf{Training and Inference Details.}
We use a C-class cross-entropy loss as the objective of our classifier. The whole network is optimized with Adam~\cite{adam} optimizer and all the parameters are initialized with values sampled from a zero-mean Gaussian distribution. We augment our data by adding a random offset to the point cloud, but still keep all the points within the gripper closing area.

\section{Grasp Candidate Generation}
\label{sec:gpg}
To build a complete grasp pipeline, grasp candidate generation is needed as a prerequisite of grasp planning.
We adopt GPG~\cite{gpg} to perform heuristic grasps sampling from the point cloud. Additionally, we propose several modifications to the original GPG to reduce the collisions between generated grasps and the support surface: 1) We discard the sampled points that are close to the support surface. 2) A grasp configuration that is approaching away from the support surface will be removed. 3) For a colliding grasp, we will try to pull it along the opposite of their approaching directions until the collision disappears. Then if there are still some points remaining in the gripper closing area, we will mark this pulled grasp as a non-colliding one.

\section{EXPERIMENT} 
We evaluate our proposed PointNetGPD both in simulation and the robotic hardware. For simulation, we mainly concentrate on the performance of grasp quality classification tasks. For experiments on the robotic hardware, we conduct several robotic grasping tasks to see whether our model can generalize well to real world settings.
\begin{table*}[ht]
\centering
\caption{Accuracy of Different Models and Configurations}
\begin{tabular}{ccccccccc}
\hlineB{2}
& \multicolumn{2}{c}{GPD (3 channels)} & \multicolumn{2}{c}{GPD (12 channels)} & \multicolumn{2}{c}{Ours (2-class)} & \multicolumn{2}{c}{Ours (3-class)} \\ \cline{2-9} 
& w/o dropout       & w/ dropout       & w/o dropout        & w/ dropout       & All classes       & Best class       & All classes    & Best class          \\ \hline
\#Params           & \multicolumn{2}{c}{3.63M}            & \multicolumn{2}{c}{3.64M}             & \multicolumn{4}{c}{\textbf{1.60M}}                                          \\ \hline
1-Viewed Point Cloud & 76.36\%           & 76.42\%          & 79.34\%            & 79.96\%          & \textbf{84.75\%}  & 86.26\%          & 79.45\%        & \textbf{90.37\%}    \\
Full Point Cloud   & 81.38\%           & 82.50\%          & 83.50\%            & 84.29\%          & \textbf{91.81\%}  & \textbf{92.18\%} & 84.15\%        & 89.76\%             \\ \hlineB{2}
\end{tabular}
\label{tab:gpd}
\end{table*}

\subsection{Simulation Experiments}\label{sec:sim_exp}
\subsubsection{Experiment Details}
In simulation experiments, we mainly want to compare the performances on grasp quality classification between our proposed PointNetGPD and current state-of-the-art methods. We choose GPD~\cite{gpd} as baseline. In the dataset, since we cannot acquire the camera location for computing the unobserved area used in the 15 channel version of GPD, we only compare the 3 and 12 channel versions (we compare with 15 channel version of GPD in robotic experiments). Also, to examine the stability on sparse point cloud, we provide either point cloud from 1-viewed or full point cloud input for each grasp. The point cloud of 1-viewed is taken from the camera in front of the object. For the full point cloud, we register the point cloud from all the available viewpoints. After the point cloud is ready, we discard the sample that has less than 50 points in the gripper closing area, then for the rest, we upsample/downsample their point cloud into 1000 points.

Finally, we run a 3-class classification experiment mainly for verifying the validity of the scores we provided in our grasp dataset. For 2-class classification, we regard a grasp with score~\eqref{label_score} above $1/0.6$ as positive, while for 3-class classification, the score thresholds for 3-class classification  will be $1/0.5$ and $1/1.2$. 

\begin{figure}[H]
    \centering
    \includegraphics[width=0.4\textwidth]{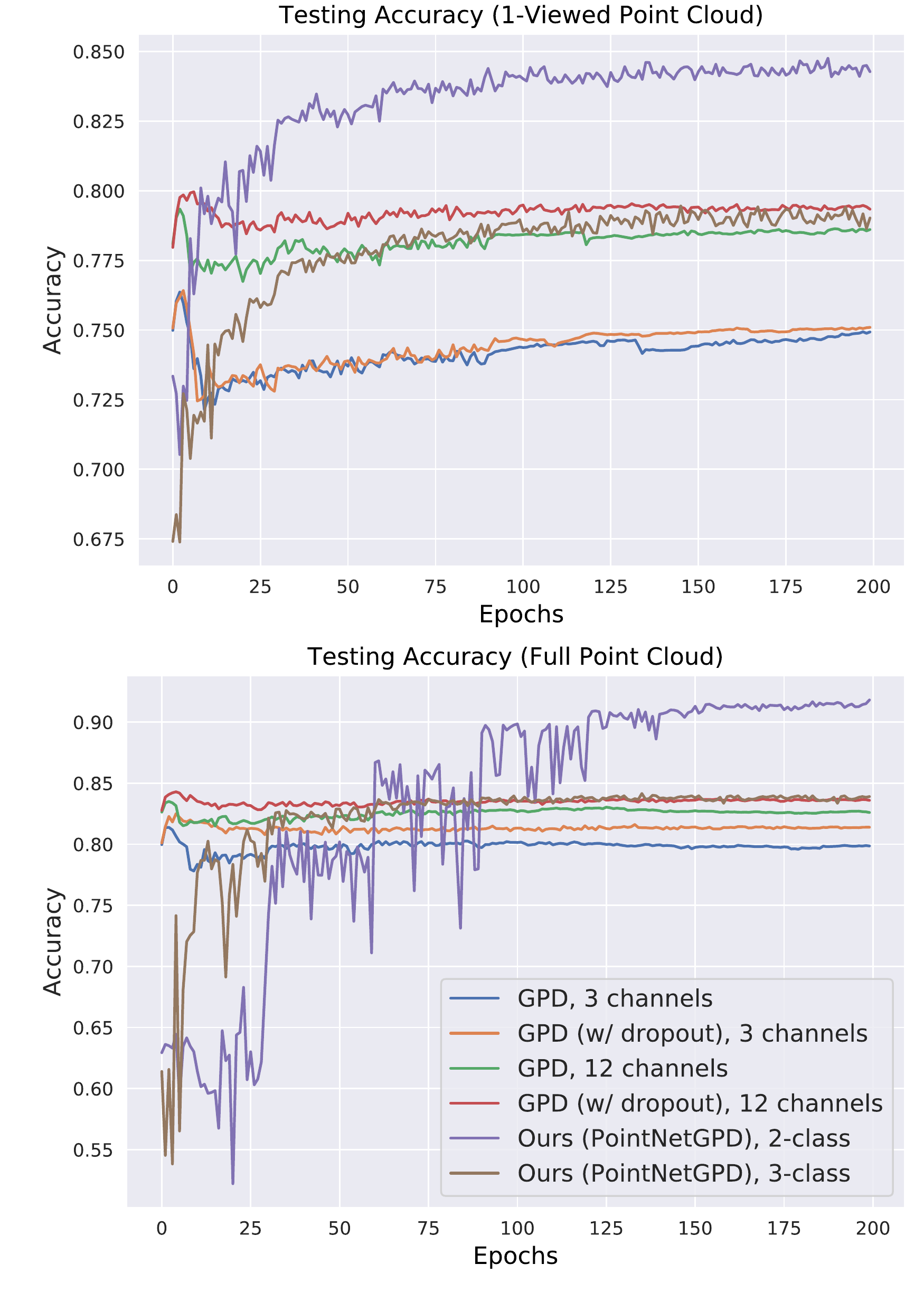}
    \caption{Classification accuracy at test set with different models and configurations on single views and on full point cloud. We can find that all the models obtain better performance with full point cloud input than with a single view, while the proposed grasp evaluation model outerperforms the baselines on both input types. More quantitative results can be found in Table~\ref{tab:gpd}.}
    \label{fig:gpd_curve}
\end{figure}
\begin{table*}[ht]
\centering
\caption{Results of single object grasping experiments}
\begin{tabular}{cccccccccccccc}
\hlineB{2}
Method & Avg. & \begin{tabular}[c]{@{}c@{}}cleanser\\ bottle\end{tabular} & mug & \begin{tabular}[c]{@{}c@{}}meat \\ can\end{tabular} & \begin{tabular}[c]{@{}c@{}}tomato \\ soup can\end{tabular} & banana & \begin{tabular}[c]{@{}c@{}}toy power \\ drill\end{tabular} & chain & \begin{tabular}[c]{@{}c@{}}mustard\\ bottle\end{tabular} & \begin{tabular}[c]{@{}c@{}}wood \\ block\end{tabular} & screwdriver \\ \hline
GPD & 49.00\% & \textbf{100.00\%} & 30.00\% & 60.00\% & 90.00\% & 20.00\% & \textbf{80.00\%} & 0.00\% & 90.00\% & \textbf{90.00\%} & 20.00\% \\ \hline
\begin{tabular}[c]{@{}c@{}}Ours \\ 2-class\end{tabular} & 81.00\% &\textbf{ 100.00\%} & 50.00\% & \textbf{80.00\%} & \textbf{100.00\%} & \textbf{90.00\%} & 70.00\% & \textbf{60.00\%} & \textbf{100.00\%} & \textbf{90.00\%} & 70.00\% \\ 
\begin{tabular}[c]{@{}c@{}}Ours\\ 3-class\end{tabular} & \textbf{82.00}\% & 90.00\% & \textbf{70.00\%} & 70.00\% & \textbf{100.00\%} & \textbf{90.00\%} & \textbf{80.00\%} & \textbf{60.00\%} & 90.00\% & \textbf{90.00\%} & \textbf{80.00\%} \\ \hlineB{2}
\end{tabular}
\label{tab:single_grasp}
\end{table*}

\begin{table*}[ht]
\centering
\caption{Results of clutter removal experiments}
\begin{tabular}{cccccccccc}
\hlineB{2}
\multirow{2}{*}{} & \multicolumn{2}{c}{GPD}  & \multicolumn{2}{c}{Ours 2-class} & \multicolumn{2}{c}{Ours 3-class (best class)} & \multicolumn{2}{c}{Ours 3-class (second class)} \\ \cline{2-9} 
& Success rate & Completion rate & Success rate      & Completion rate     & Success rate             & Completion rate             & Success rate              & Completion rate              \\ \hline
\multicolumn{1}{c}{Set 1} & 84.83\%     & 95.00\%           & 86.54\%          & 94.08\%            & \textbf{89.33\%}                 & \textbf{100.00\%}                      & 52.10\%                  & \textbf{100\%}                       \\
\multicolumn{1}{c}{Set 2} & 61.13\%     & 81.50\%         & 61.07\%          & 84.38\%            & \textbf{66.20}\%                 & \textbf{95.00\%}                       & 43.75\%                  & 37.50\%                        \\ \hlineB{2}
\end{tabular}
\label{tab:clutter_grasp}
\end{table*}

\subsubsection{Results Analysis}
The testing accuracy of all the considered models during training is demonstrated in Figure~\ref{fig:gpd_curve}. We list the best result among the 200 epochs in Table~\ref{tab:gpd}. Here we highlight some important facts we found in these results. First, our proposed PointNetGPD performs significantly better on grasp quality classification than all the GPD baselines. Even on the most difficult 1-viewed point cloud, PointNetGPD still has an averaged $4.79\%$ improvement over the best GPD baseline. Furthermore, from Figure~\ref{fig:gpd_curve} we can see that GPD can easily get overfitting on the training set. However, although we make it easier by utilizing Dropout~\cite{dropout} on the GPD network, there is still a performance gap between GPD and our method. Such results are partly due to the number of parameters. Compared to GPD, the network in our proposed method has fewer parameters and performs better, which means that our network is more effective regarding geometry analysis especially from the sparse point cloud. 

For the 3-class experiment, we found that the accuracy of the class of the best quality is even better than the best class in the 2-class experiment. This may imply that a grasp with higher score~\eqref{label_score} will be easier to identify. In our robotic experiments, we will make further validations by comparing the results of 2-class and 3-class grasping models.

\subsection{Robotic Experiments}\label{sec:real_exp}
We validate the reliability and efficiency of our proposed PointNetGPD in two robotic experimental conditions: objects were presented to the robot in isolation as well as in a clutter. 
These experiments were carried out on a UR5 robotic arm with an attached Robotiq 3-finger adaptive robot gripper. As shown in Figure \ref{fig:robot_setup}, the gripper works under pinch mode, in which only two contact surfaces are allowed to move toward and away from each other along a 1-D manifold. Especially, since we only use one Kinect2 depth sensor, all the point cloud provided in robotic experiments is 1-viewed, which makes it even more challenging.

We select 22 objects from the YCB object set. In these objects, 11 of them have already been presented in our grasp dataset, while the rest are novel. We also select 16 from 22 objects to construct two object sets that used for clutter removal. Details can be found in Figure~\ref{fig:objects}.

The whole system is implemented using the ROS framework, 
particularly, a fast hybrid evolutionary inverse kinematics solver BioIK \cite{bioik} is used for solving inverse kinematics within the MoveIt! framework.

For both conditions, we compare a 2-class and a 3-class PointNetGPD with a 15-channel GPD baseline. In addition, to validate the significance of the quality scores provided in our dataset, we also compare the grasp performance between the best and the second class in 3-class PointNetGPD.

\subsubsection{Objects Presented in Isolation}
In this experimental condition, all the objects presented in Figure \ref{fig:objects} are tested.
We test each object for ten rounds with random initial orientations.
If the gripper failed to grasp an object or no collision-free grasp pose was generated within a long time (in our practice we use 5 minutes), we mark this attempt as failed. We only consider the success rate for performance evaluation in this experiment.

Table \ref{tab:single_grasp} demonstrates the grasping results for a single object using three different models. Note that Table \ref{tab:single_grasp} does not contain the objects whose success rates are 100\% for all the three models, such as chips can, Rubik's Cube, plastic apple and so on, or 0\% such as the medium clamp. The 0\% success rate of this object is probably caused by the poor quality and the low height of the acquired point cloud, and the irregular shape. 
As Table \ref{tab:single_grasp} illustrated, the two types of PointNetGPD methods manifest a higher average success rate, which suggests that the proposed model can better understand the spatial geometry of the point cloud in the graspable region. 

\subsubsection{Objects Presented in Dense Clutter}
\begin{figure}[H]
    \centering
     \subfigure[]{\includegraphics[height=0.2\textwidth]{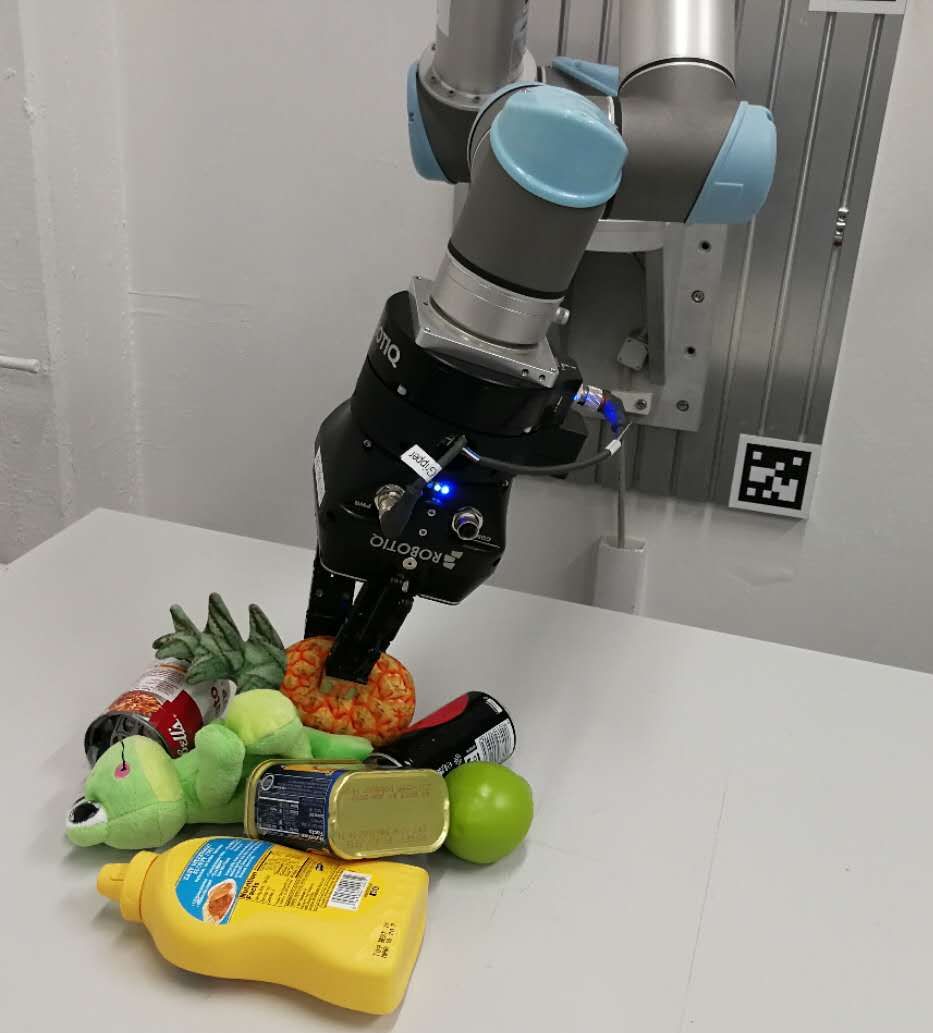}\label{fig:robot_setup}}
    \hspace{0.1in}
    \subfigure[]{\includegraphics[height=0.2\textwidth]{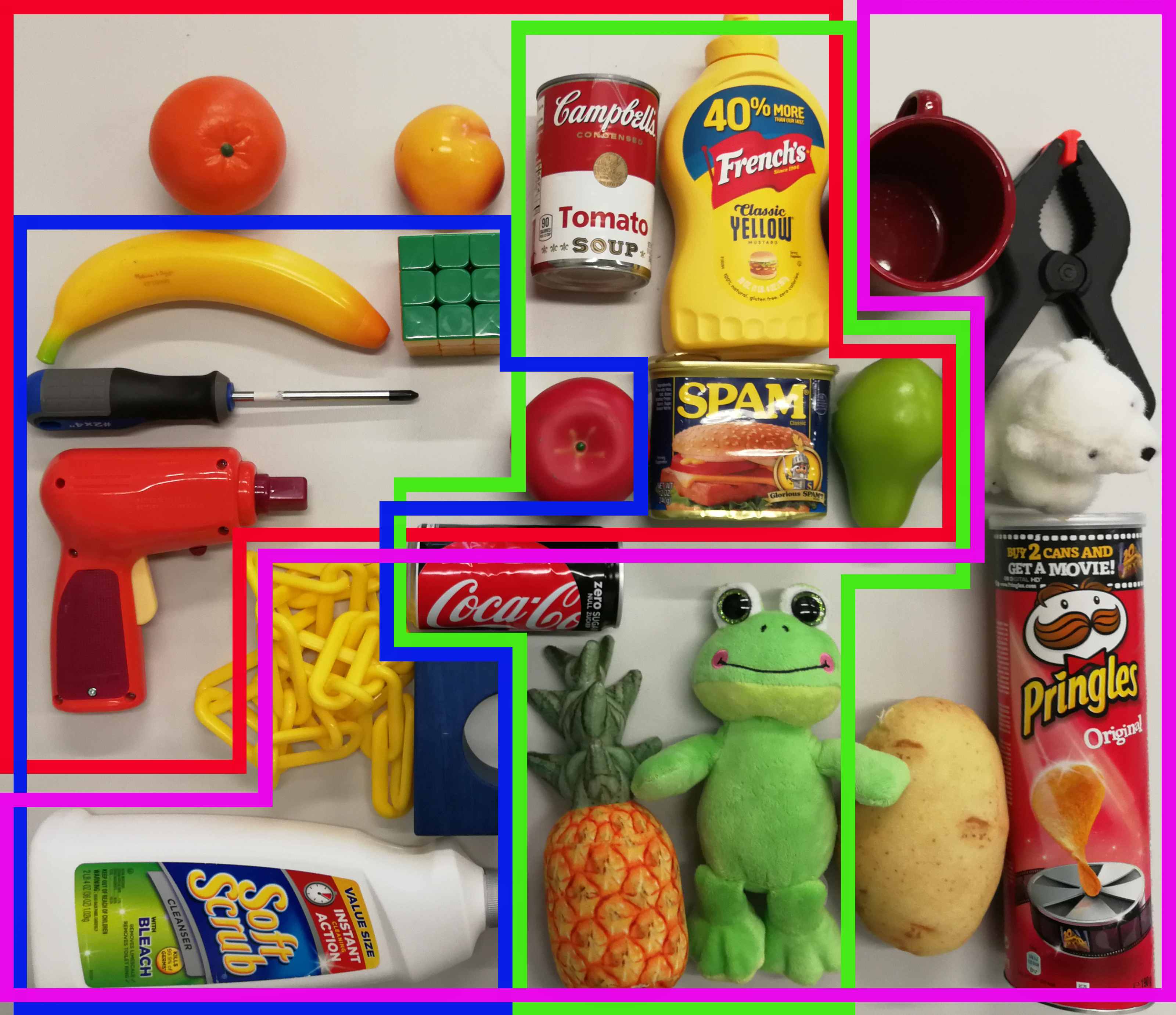}\label{fig:objects}}
    \caption{Settings of our robotic experiments. \subref{fig:robot_setup}~Grasping experiment setup with UR5 robotic arm and Robotiq 3-finger adaptive robot gripper. \subref{fig:objects}~Objects used in our experiments. Red polygon shows the objects presented in the training dataset, magenta polygon contains the objects that are not in the dataset. The green (clutter 1) and blue (clutter 2) polygons present the two object sets used in clutter experiments respectively.}
    \label{fig:robot_two_pic}
\end{figure}

In dense clutter condition, we select 16 objects from those who have grasping success rate above 0 for all the compared model to construct two object sets (Set 1 and Set 2). The green and blue polygons in Figure~\ref{fig:objects} represent these two sets respectively. Furthermore, Set 1 has six objects with 100\% success rate in isolated condition for all the three models, while Set 2 only have two objects with 100\% success rate. We run experiments with each object set for five rounds. 

Besides the models we compared in the isolation condition, here we also test the grasp that is predicted to be the second class through our 3-class PointNetGPD. This is mainly to verify the validity of the multi-class classification. We use \textit{success rate} and \textit{completion rate} as the criterion for performance evaluation. The \textit{success rate} is the percentage of successful grasps, while the \textit{completion rate} is the percentage of objects that are removed from the clutter.

From the results presented in Table \ref{tab:clutter_grasp}, we found that all the models overall perform better in Set 1 than Set 2 because the objects in Set 1 could better fit the geometry shape of the gripper, and some of them have a higher roughness. 
Meanwhile, grasps from the best class of 3-class PointNetGPD show the best grasping outcomes, especially on \textit{completion rate}. This shows a significant averaged improvement of $13.5\%$ over GPD. Moreover, the fact that grasps from the best class of 3-class PointNetGPD are hugely superior to the second class confirms the effectiveness of 3-class classification, which implies the capability and implication of the meticulous scores in our dataset. 

Occlusion could cause failures in this experiment since we only have one fixed view of the point cloud. Additionally, sometimes the model may treat multiple objects as a single one and attempt to grasp them together, and this could also induce failures.

\section{Conclusion and Future Work} 
We have presented PointNetGPD, a novel approach for detecting grasp configurations from point sets. As the core module in our grasp pipeline, we proposed to address the challenging grasp quality evaluation over imprecise and deficient point cloud with PointNet~\cite{pointnet}. To further improve the performances, we generate a large-scale grasp dataset with 350k real point cloud and grasps with the YCB~\cite{YCB} object set for training. Our experiments show that our model outperforms the state-of-the-art grasp detection methods. 

In future work, our goal is to integrate the grasp candidate generation step into the network for performing grasp planning in an end-to-end fashion. Additionally, we plan to do clutter segmentation simultaneously, which can prevent the model from planning unfeasible grasps that cross more than one object.

\small{
\section*{ACKNOWLEDGMENT}
This research was funded jointly by the National Science Foundation of China (NSFC) and the German Research Foundation (DFG) in project Cross Modal Learning, NSFC 61621136008/DFG TRR-169. It was also partially supported by National Science Foundation of China (Grant No.91848206, U1613212) and project STEP2DYNA (691154). We would like to thank Chao Yang and Professor Huaping Liu for their generous help and insightful advice. 
}
\bibliographystyle{IEEEtran} 
\bibliography{IEEEabrv,ref}

\end{document}